# A Controlled Set-Up Experiment to Establish Personalized Baselines for Real-Life Emotion Recognition

Varvara Kollia, Noureddine Tayebi

{Varvara.Kollia, Noureddine.Tayebi}@intel.com[1]

**Abstract**
We design, conduct and present the results of a highly personalized baseline emotion recognition experiment, which aims to set reliable ground-truth estimates for the subject's emotional state for real-life prediction under similar conditions using a small number of physiological sensors. We also propose an adaptive stimuli-selection mechanism that would use the user's feedback as guide for future stimuli selection in the controlled-setup experiment and generate optimal ground-truth personalized sessions systematically. Initial results are very promising (85% accuracy) and variable importance analysis shows that only a few features, which are easy-to-implement in portable devices, would suffice to predict the subject's emotional state.

1. Introduction
Recent advances in sensor technology open new possibilities in all aspects of human-machine interaction. One of the new areas of interest is the understanding of human cognitive and emotional behavior through bio-signals. Miniaturized physiological sensors allow for seamless integration in portable devices and lend themselves to providing continuous insights into the user's emotional state. The interpretation of these signals into distinct emotions is possible through pattern matching, via a machine learning algorithm.
Typical physiological sensors used for emotion recognition are: electrocardiogram (ECG), electromyogram (EMG), body-temperature (T), galvanic skin response (GSR), photoplethysmogram (PPG) and even brain waves by electroencephalogram (EEG) sensors. It has been demonstrated that there is clear correlation between the emotional states of the subject and these sensor-data. However, the application of trained models into real-life has always been challenging, with the main difficulties stemming primarily from the fact that there is no unique definition of emotions, as well as the subjective nature of the problem and the variety of other-factors that need to be taken into consideration, when applying controlled-setup models to real-life.
In this work, we tackle the emotion recognition problem at a personalized level, to take into account its subjectivity. In addition, we treat emotions largely as states, rather than instances, to be able to characterize more reliably the sensor data and the emotional states themselves are defined to be clear, distinct and easy to stimulate, based on the available material. The controlled setup-experiment itself is highly personalized. The final goal of this would be to employ the personalized models at similar static real-life conditions and improve them subsequently, based on the subject's feedback.

---

[1] *The main part of this work took place in Q2-Q3, 2015.  Corresponding affiliations refer to that period.*

## 2. Problem Definition

The main target of this study is to set guidelines for establishing the ground-truth for personalized models in emotion recognition problems using biosignals. In particular, we are trying to answer the question of telling one's emotional state from biosignals, by breaking down it down to the following sub-problems: i) Ground-truth problem: training a model on predicting the emotional state of the subject under constant external conditions in a controlled set-up environment where the provoked emotion is known a-priori, to set the baseline and ii) Real-life prediction: repeating the experiment under similar conditions with real-life data from the subject to predict their emotional state in a summarizing fashion, and compare to (any) input from the subject, if available, using the trained baseline model of the first stage. Due to the subjective nature of the emotion recognition problem, we opt for a personalized model, where separate baseline experiments, with different stimuli, will be used to set the baseline for each subject at a personalized level.

More specifically, the goal of the experiment is to establish the ground-truth for six emotions at a personalized level. The user rankings are used as guidelines, as the experiment progresses, for future stimuli (clip) selection, as well as for data processing. The main criterion for clip-selection is to provoke strong intensity clear emotions that can form a baseline. A secondary goal is to rank the effectiveness of different bio-signals in emotion classification. Throughout this study, we will be using easy-to-derive features in time domain, ideal to be implemented in wearable devices. Also, the experiment is conducted with a more general goal in mind, to establish the baseline for the subject so that we are able to predict their emotions (in general categories) in real-life situations in the future, under similar (static) conditions, e.g. while watching movies, doing office-work, during a social/professional visit or while being engaged in activities that would not require significant movement/exercise. If this study were to be used as emotion-baseline for other aspects of the subject's life, a systematic way of evaluating the effect of movement, as well as context information (eg environmental factors) would need to be in place. The baseline model can be extended to more complex environments and settings, once the additional factors we need to consider are evaluated systematically through the use of additional sensors, and other corresponding sources of relevant information, such as other mobile/portable devices.

### 2.a Sensors

A schematic of the problem can be seen in Fig. 1. As input signals, we will use the Heart-Rate (HR) signal and its derivatives , namely heart rate variability (HRV), heart rate percent (HRP) and breathing rate (BR), as provided by electrocardiogram (ECG) and/or photoplethysmogram (PPG) sensors, as well as Galvanic Skin Response (GSR) and Skin Temperature (SKT). Brain waves and muscular movements, as recorded by electroencephalogram (EEG) and electromyogram (EMG) sensors were also considered, but will not be part of this study, to maintain the recording system's ease-of-use and portability. Once these signals are recorded, they will be processed with machine learning methods to provide the estimate for the subject's emotional state.

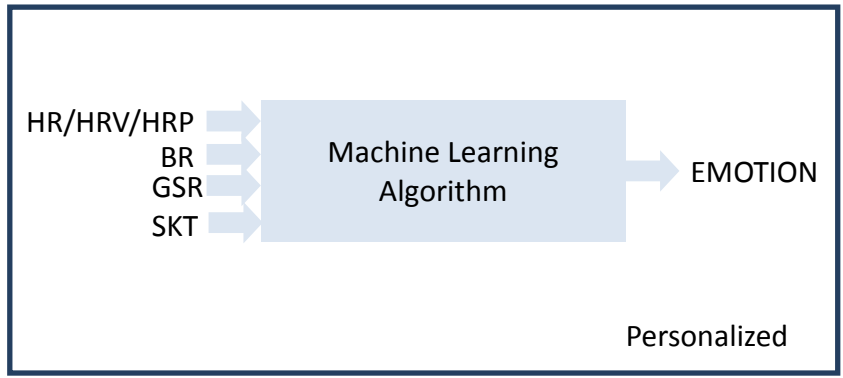

Fig. 1. System for Personalized Emotion Recognition.

Two collection devices were used for this experiment: the Zephyr bioharness was used to report HR (and its derivatives) from ECG sensor, and an in-house device was used to provide GSR (from the fingers) and SKT (from the wrist). ECG and PPG sensors for HR recording were also available from our recording device. The Zephyr bioharness is placed across the chest and an interface is available for signal collection, pre-processing, feature extraction, and storage. It is noteworthy, that as two separate devices were used, special attention was necessary for their time-alignment and interpolation/extrapolation to handle missing values and correct for different sampling rates.

**2.b Targeted Emotions**

We will try to identify three positive and three negative distinct emotions. The six emotions we will be reporting on this experiment are the following: sadness/anger, fear, disgust, awe/reverence, contentment, joy/amusement.

The definitions we use for the three negative emotions are as follows. Sadness is considered the primary emotion, anger may or may not be elicited as secondary emotion, or as a means to cover the primary feeling which is sadness. Sadness is the emotion that provokes reaction of tears, nervousness and helplessness. It makes one feel unhappy, miserable, desolate and in pain. Fear is the feeling that one is in danger and that there is an imminent threat. This feeling may cause panic and state of high alert. Disgust is the repugnance caused by something hard to look (or sense in general), which is repulsive and horrid.

The three positive emotions are awe/reverence, contentment and joy/amusement. Awe/reverence is the feeling associated with admiration at the sight of a wonder, and it includes the element of surprise; it is used in the sense of admiration for something exceptional. In contrast, contentment refers to a state of relaxation and happiness, it is the state of calm satisfaction. Joy/amusement is used here in conjunction to laughter; it is the emotion associated with something that one finds entertaining.

These categories were chosen in terms of distinctiveness and relative ease of stimulation, in order for the experiment to be effective. By no means, is this a complete definition of emotions. These emotions were not selected to cover the entire emotional spectrum of the individual, but rather as a clean subset that lends itself to identification, with distinct boundaries among its members. That is the reasoning behind grouping anger with sadness, since it is usually hard to clearly separate the two. Likewise, as joy is a very general category, we target specifically joyful emotions related to amusement (laughter) only. We should also note that a small set of targeted emotions should always be defined beforehand, in order to provide a starting point for material selection and experiment definition. However, the definition is not hard and it should be adjustable to some degree to the individual; mostly towards the direction of merging (or excluding) categories, than towards the one of greater resolution, to allow for greater accuracy. Therefore, the emotions for the current study were not chosen to cover the entire range of emotion of (any) subject, but they were more meant to be used as broad categories, that do not have much overlap and can be stimulated clearly.

## 3. Experiment

In this section, we describe the experiment design, the stimulus selection, ranking and the protocol. We also touch on how the user's feedback is currently incorporated into the experimental design and how this process can be automated and optimized to allow for more robust emotional baseline generation. We note that even though this experiment was designed for and conducted with adults, it could be used to establish baselines on children, in order to extract clear distinct emotions, by adjusting the stimuli selection and the emotion definition appropriately. The reasoning behind this is that children usually experience clear emotions of strong intensity of duration that is easily distinguishable. A direct real-life application where the trained baseline model could be tested would be in school, during hours of teaching and performance evaluations.

### 3.a Stimuli Selection

Publicly available video-clips are used for the most part, in this experiment. There are two main reasons behind this choice: i) the availability of a large number of online selections from different categories and ii) the combination of visual and hearing elements which would lead to greater efficiency in evoking particular feelings. More specifically, the stimuli (clips) are carefully selected from publicly available *youtube* clips. In order to be appealing to a greater audience, clips relatively popular were chosen, that gathered positive feedback. However, as the element of surprise is typically important, the most popular clips were left out on purpose, even though we generally aimed at widely accepted material, in terms of its efficiency to stimulate the desired response, based on other users' comments. Clips are selected from the news, from the movies and from documentaries/tv-series for the most part. Likewise, we aim at selecting relatively contemporary clips mostly; to maximize their effectiveness, keeping in mind that it is better if the subject has not been exposed to them for a while. In the long run, all this manual work should be substituted by a smart automatic selection system that would be based on an initial questionnaire and the user's feedback as the experiment progresses.

In evoking the negative emotions, there are some generally accepted historical events which are tragic, that are bound to result in strong sadness for most people. These are used as the main pool of sadness and anger stimuli. Obviously, factors such as the age of the subject and their origin are, among others and, to some extent, factors that should be considered in further personalizing the material. With respect to fear, well-known scenes aiming at provoking fear from the movies are selected, as well as publicly available user-compilations of relevant clips. These scenes should be independent and complete, in the sense, that they should bring a person of neutral state to a person in the state of the targeted emotion, without any other background/information necessary. Finally, there are many clips based on personal experiences, as well as scenes from movies and TV-shows targeting disgust.

It is noteworthy, that negative emotions, especially sadness/anger and fear take time to build-up, therefore very short clips should be avoided. Additionally, very large clips, as the subject gets tired, may lose in intensity and should, therefore, not be considered. We aim more at characterizing biosignals when the subject is at a state of e.g. fear, than at capturing strong brief moments of fear, as the second case would be much harder to identify consistently.  Specifically, with respect to the negative feelings, no personal material should be selected, even though personalization is possible; e.g. there are clips which would provoke extremely negative feelings, e.g. certain news stories, which would be selected based on previous input by the user.

On the other hand, there are some generally accepted incredible talents and acts/performances or places also bound to provoke admiration (awe) or other positive feelings. High-ranked funny compilations in terms of everyday events and short movie-parts from various popular sources are used to evoke joy, which is here used in terms of amusement. Note that long sessions in terms of positive feelings should be fine; it is mainly the negative feelings that require interchanging. Finally, calm, serene clips (e.g. from nature), are used to help the subject experience the contentment feeling.  Music can also be a very important element in these clips in choosing among similar stimuli of the same category. In this manner, in designing the experiment we can take into account some general guidelines in selecting our material that would be then further personalized based on the subject.

Special attention should be drawn to the fact that the user's input is very important for all these selections. For example, someone's specific sensitivities (or lack of) and general beliefs on certain categories, could guide the stimuli selection appropriately.   For example, if someone perceives a certain category of clips aimed at provoking fear as funny, these clips should be clearly avoided when targeting fear, in the next sessions.

With respect to the fully personalized session, someone close to the subject should pick the stimuli, in order for the subject not to be exposed to the material beforehand and fully experience the emotions during the session. The stimuli can be personal, by e.g. including family videos not seen for a while, or it can be merely personalized, by selecting material of special significance to the subject, which is not personal, (e.g. scenes from their favorite comedy).

The stimulus-pool should be subject to a few fundamental rules that would make it applicable to most people.  In this work, emphasis is placed in finding effective *youtube* videoclips that would stimulate the targeted emotions, whereas no hard limits are imposed on the total duration of each emotion/session, as long as it falls within a reasonable window (60-70min), to facilitate the clip selection. If we have more data for some emotions, that would let us take-out clips or sessions that received lower rankings. On the other hand, if some sessions receive low scores, more data can be collected to replace those sessions/ emotions as the experiment progresses. Ideally, the subject's baseline will only be constructed from (parts of) sessions that actually provoked the desired emotions and any unreliable data will be taken out.

The length of the clips is subjective, however, in general, the clips should neither be too long nor too short; 5-10min clips are usually ideal, except in the case where the clips are too strong, in which case ~3min should suffice to evoke the targeted emotion. In the case of very short clips, compilations should be used.  We used 10-15min compilations of short clips for the positive feelings mostly, as compilations that consist of short entertaining or awe-inspiring clips would get one's focus and stimulate their curiosity on what comes next. Longer clips lose on the intensity of the provoked emotion. In general, each session should contain 2-3 emotions, with smooth transitions. It is also reasonable to expect that most people would prefer to transition from negative to positive feelings and close on a good note. Experiencing the negative emotions is harder, therefore sessions with predominantly negative emotions should be alternating between emotions and possibly be shorter in total duration. Full sessions of one positive emotion (eg. joy) may not be problematic, as long as they keep the subject's interest.
In this particular experiment, we found that the subject had a preference for short clips that escalate. Additionally, the most provocative clips got the best rankings and clips that included music had a great effect on the subject. Finally, the fully personalized session got consistently high rankings.

### 3.b Stimuli-Ranking

The user should give their feedback at the end of each complete session, to avoid loss of focus. The interviewing process is kept simple and clear. The main feedback the user provides is stimuli-ranking; each clip is ranked on a scale of 1.10; 1 being the lowest and 10 being the highest score, with respect to its effectiveness in provoking the targeted emotion. For example, if one clip is aimed at provoking sadness/anger and the user ranks it with 10, this translates to the clip being very efficient in making the subject feeling really sad or angry. Other reporting techniques are also available and will be considered in the future, however, direct ranking helps keep the reporting simple and effective. In addition, the subject can provide feedback on which emotion(s) did the clip actually provoke, in case they are different from the desired one. The user can additionally report on the length of each clip and which parts in particular were most effective (for the longer ones).  This feedback is used to select future clips that would stimulate the desired emotions on the subject more efficiently.

### 3.c Experiment-Protocol

In this section, we will briefly describe the experiment's protocol. A general protocol is established, with a certain degree of personalization, as it will be explained shortly. The protocol consists of nine sessions in total, eight sessions ~60min each and one session which is fully personalized, with 10-15 clips/session on average. The order of stimulated emotions is interchanging in each session.  The order of emotions in each session is of importance; abrupt transitions were avoided as much as possible, (e.g. from joy to sadness), to facilitate the actual emotional experience of the subject.  The last session is personalized, based on the subject's preferences; the material itself does not have to be personal (though it could be), but it has to reflect the subject's ideas/tastes on what they think is most likely to provoke them the feeling in question. The number of emotions in each hour-long session should be limited to 3, otherwise the intensity of the stimulated emotion may be compromised. It is also noteworthy that in this experiment 5min resting-periods were used to separate different emotions. This should be adjustable based on the user's feedback; here, it was found that 3min should suffice.
Regarding the first eight sessions, they are based on clips drawn from the general material, and selected according to the user's feedback.  Each one of the eight sessions contains 2-3 emotions and it lasts ~60-70min, on average. The sessions start with ~5min of rest (calmness) and the emotions of each session are also separated by ~5min of rest. Therefore, each session contains ~15-25min of distinct emotions, in most cases. The duration of the clips varies with the majority being ~3-12min; however we also included compilations with very short clips (~1min) and total duration ~40min. The compilations of short clips target positive feelings, (joy and awe/reverence); as the negative emotions need longer time (clips) to build up. A typical breakdown of a session would contain 2-3 emotions, each of which would contain 2-4 clips; with each emotion being separated from the next one with a period of rest. The last session is fully personalized and it consists of ~10min of personalized clip selections, as provided by a person close to the subject. The order of emotions is as follows: sadness/anger→fear→disgust→awe/reverence→ contentment→joy/amusement, separated by 3min resting periods. This order was chosen to allow for as smooth transitions as possible, and close on a good note, and it was found to be effective. The personalized session can be broken into 2 sessions, if the number of emotions is found to pose a problem.

A breakdown of the experiment can be seen in Table I. In total, the experiment took ~9.1 hrs, with 3.8 hrs of positive (awe/reverence, joy/amusement, contentment) and 3.2 hrs of negative (fear, disgust, sadnesss/anger) emotions, and the remaining being periods of calmness (rest-in-between).

We have a total of 1027 instances, using a window of w=32 samples, at a sampling frequency of $f_s$=1Hz; after undersampling. Note that the HRV/BR signals we use, were already extracted at a much higher frequency. In terms of duration, we collected 192 min (3.2 hrs) of negative emotions and 228 min (3.8 hrs) of positive emotions. The remaining 2.1 hrs were resting-periods (neutral), which will not be included in the classification, due to the fact that they included transition periods between emotions, as well as some motion artifacts. We should also note that the highest-ranked data of the fourth emotion (disgust) were not recorded, due to hardware failure; which is the reason for the relatively low number of instances of this emotion. This controlled setup experiment is a long experiment; most controlled setup experiments are ~ 1hr, and it was demanding on two levels: it required a lot of carefully selected material and it was emotionally draining, due to the very strong material, as well as to the subject's focus and effort. On the other hand, it is a reliable way of setting the ground-truth with a limited number of inputs for real-life scenarios, and its shortcomings can be overcome with a smart automated system.

Table I: Summary of Experiment

| Label | Symbol | Frequency | Duration (min) |
|---|---|---|---|
| Rest | 0 | 240 | 128.0 |
| Fear | 1 | 129 | 68.8 |
| Sad/Anger | 2 | 122 | 65.1 |
| Awe/Rev | 3 | 158 | 84.3 |
| Disgust | 4 | 109 | 58.1 |
| Joy/Amus | 5 | 149 | 79.5 |
| Content | 6 | 120 | 64.0 |

It is also noteworthy that it was found that the particular mood of the day matters and cannot be fully removed with signal normalization. Ideally, the sessions should be provided in the beginning of the day, when the subject is alert, responsive and able to focus. As the day progresses, it is easier to lose focus, respond less to the positive feelings and small destructions can be detrimental for the experiment. Further quantification on how the particular mood (as related to the time of the day) affects the results is beyond the scope of this study and would make the experiment longer and more demanding. It is better to shorten the experiment by creating an effective adaptive selection system and offering the sessions in the beginning of the day.

**3.d Protocol Extension for Automated Personalized Selection**
The main target of this study is to set guidelines for establishing the ground-truth for personalized models in emotion recognition problems. The user-rankings and feedback are used as guidelines to select the material of future sessions. E.g., if the targeted emotion is fear and the subject thinks that supernatural clips are funny, then supernatural clips should be excluded from future sessions in the controlled set-up experiment to get an accurate baseline for this emotion. This adaptive material selection is novel and even though, it was presently done manually, an automated system can be built to adjust the material-selection and re-define the future sessions of the experiment in an automated fashion, by e.g. using clustering to group stimuli (clips) based on key-words and recommend material for the specific user based on their so-far feedback.

A pre-requisite for this automated smart session/experiment-generating system would be the existence of a large pool of emotion-stimulating material. The selection mechanism should make its recommendations from a large pool of initial stimuli and be adaptive to generate the most effective subset, to establish the baseline for the subject's emotional response. The initial stimuli would be created by a number of different people of different ages and backgrounds. It should also be mentioned that the order of the sessions can change, depending on the particular mood of the subject. However, conducting the experiments at a particular time of day under regular conditions should help with setting unbiased baselines and generalization.

Based on the same idea of personalization by utilizing the user's feedback, our manual process can be automated (and shortened) via a smart selection system. Once a large pool of stimulating material is formed based on certain ground rules, some of which have been described above, the first sessions can be formed by selecting material according to the subject's responses on an initial questionnaire. The questionnaire should provide general personalized guidelines based on the user's answers regarding their general preferences and ideas on what categories/subjects of the pre-selected material may cause them the emotions in question. The following sessions should be based on the user's feedback and (so-far) rankings. The process should be repeated until a maximum number of iterations has been reached, or until convergence (i.e. good rankings) have been collected. This is illustrated in Fig. 2.

In the system of Fig. 2, not only rankings of previous sessions are used as guidelines for future choices, but also the targeted emotions can be adjustable based on the user's feedback, the available material and the experiment's progress. Eventually, the smart stimuli selection would make the experiment short (in terms of duration) and accurate, so that the baseline can be used to predict the subject's response in similar settings in real-life.

We should note that personalization should not lead to the development of a full-recommendation system, given that existing categorizations can be used; e.g. in terms of movie-genres, so that the initial selection of stimuli depends on the selection of certain broad categories, based on a (tree-structured) questionnaire and it should be adaptable with the course of the experiment. Also, pre-ranked clips and (e.g.) *youtube* recommendations can be utilized in personalizing the material further. The adaptive model will learn further and be improved based on the user's feedback in real-life, but the controlled-setup experiment should provide a good starting-point in establishing solid ground-truth baselines.

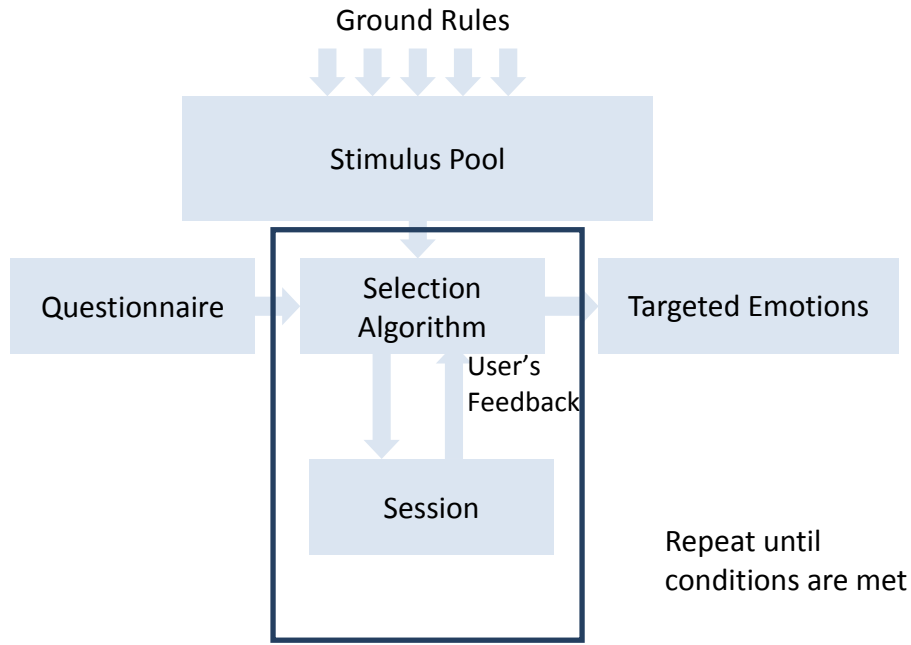

Fig.2 Smart stimuli selection system to allow for personalized ground-truth detection.

## 4. Data Analysis

In this section, we will present the data analysis methodology and discuss in detail the results of the experiment. We also rank the features in terms of their effect in classification accuracy, both for the six-emotion, as well as for the binary setup. Prototype code in *Matlab* and *R* was used for this analysis.

### 4.a Pre-processing

Signal pre-processing is essential to formulate our problem mathematically. In our case, signal pre-processing consists mainly of time-alignment and normalization. We reference all our measurements to a global clock, to allow for a single reference with respect to features and labeling. Additionally, interpolation and extrapolation (to handle missing values) was necessary, for uniform reporting. Our final features-report was generated with frequency 1Hz (after sub-sampling). In the pre-classification stage, the data (features with labels) are randomized after being generated from normalized signals. Signal normalization was necessary to take out the bias of the specific day/time when the session was conducted. Note that some classifiers require that features (not only signals) are also normalized, especially if feature normalization is not included in their (default) implementation. Before feature-extraction and labeling, synchronization was required for all measured signals. With respect to the GSR data, the signal was filtered with a median filter of order 10 for smoothing and for noise removal. Moreover, ECG collected data from the external device were found to be more accurate (from the PPG recorded ones) and were used in the analysis below. Finally, it was observed that the skin temperature was following a specific pattern in each session, which led to a target-leak in the system, therefore that signal had to be removed.

### 4.b Feature Extraction

We extract the features (based on the pre-processed signals prior to undersampling) shown in Table II. A window of 32 sec is applied beforehand, which, after some investigation, was found to be efficient for algorithm robustness and generalization purposes. This will be illustrated in the Results section.

Table II: Features

| Signal | Feature |
|---|---|
| HRV | Mean |
| HRV | St. Deviation |
| BR | Mean |
| BR | St. Deviation |
| HRP | Mean |
| HRP | St. Deviation |
| BR | Sum of Sq. |
| GSR | Sum of Sq. |
| HRV | Mean of Diff. |
| HRV | St. Dev. of Diff. |
| GSR | Mean |
| GSR | St. Deviation |
| SKT | Mean |
| HRV | Mean of Diff. Sq. |
| HRV | Std. of Diff. Sq. |
| HR | Mean |
| HR | St. Deviation |

The features consist of the mean and standard deviation of the main collected signals; namely HR and its derivates HRV and HRP, as well as BR and GSR. The mean of SKT, and the mean and the standard deviation of the squares of successive differences of HRV and the sum of squares (power) of BR and GSR are also extracted. In total, 17 time-domain features are extracted, which are easy to implement in wearable devices.

**4.c Results**
We use random forests, as our main classifier and present the results for both the six-emotion setup, as well as the binary setup, which refers to positive vs negative emotions classification. Random forest was found to be the most effective classifier, among the ones we tried. We also present a ranking of the relevant importance of the features.

**4.c.i Random Forest Classification**
In this section, we present the classification results with random forests (RF), which is an ensemble classifier. The results are presented for both the binary and the six-emotion setup in terms of the prediction error, which is the out-of-bag (OOB) error in the case of RF. Throughout our analysis, we report the best observed OOB error, using RF. However, it is noteworthy that we did not observe any significant OOB variability (~3%) after multiple runs from different seeds.

We start our analysis by modifying our initial setup, to exclude SKT, as input. The reasoning behind this, is that even though we found that SKT ranked high on variable importance, its behavior was similar regardless of the session, (increasing with time). This cannot be meaningful, as each session contains different emotions (in different order) and it can introduce bias if patterns are not clear enough. Therefore, to prevent a target-leak and unrealistically optimistic results, we excluded SKT from our analysis. The effect is clear in Table III, where the OOB is reported with and without SKT. Specifically, we get ~36% OOB error excluding the SKT information, whereas we get only ~27% OOB if we include it, for the 6-emotion problem. An improvement of ~7% is also observed for the binary case. We note that the negative emotions refer to fear, sadness/anger and disgust; where positive emotions refer to awe/reverence, joy/amusement and contentment.

Table III: Effect of SKT on results from all sessions

| Label | OOB (%) with SKT | OOB(%) wo SKT |
|---|---|---|
| Fear | 32.6 | 45.7 |
| Sadness/Anger | 25.4 | 35.2 |
| Awe/Reverence | 23.4 | 31.6 |
| Disgust | 30.3 | 35.8 |
| Joy/Amusement | 20.8 | 28.9 |
| Contentment | 29.1 | 43.3 |
| Average OOB | 26.6 | 36.3 |
| Negative Emotions | 19.7 | 28.1 |
| Positive Emotions | 11.7 | 18.0 |
| Average OOB | 15.4 | 22.6 |

To improve the prediction accuracy, we will repeat the analysis, taking into account only the highest-ranking parts; that is the parts of the sessions that were most effective in provoking the targeted emotions. In Table IV, we see the break-down of the highest-ranking parts of the experiment. In particular, we aimed for a minimum rank of 7 on a 1.10 scale. We end- up with ~50min/emotion and a total of ~300min (hrs) of recorded data. It is noteworthy that all the clips from the last session, which was fully personalized were included in the analysis, as these clips received consistently high rankings.

Table IV: Experiment summary using only clips that received high rankings

| Label | Symbol | Frequency | Prev. Frequency |
|---|---|---|---|
| Rest/Excluded | 0 | 465 | 240 |
| Fear | 1 | 105 | 129 |
| Sad/Anger | 2 | 84 | 122 |
| Awe/Rev | 3 | 124 | 158 |
| Disgust | 4 | 50 | 109 |
| Joy/Amus | 5 | 125 | 149 |
| Content | 6 | 74 | 120 |

We achieve ~7.6% OOB improvement for the 6-emotion setup and ~7.3% for the binary one, by taking into account only the highest-ranked clips, as we can see in Table V and Table VI, respectively. The performance could be further improved if we could tighten the ranking threshold, i.e. by taking into account only clips ranked 9 or 10. This is not feasible here, due to the fact that the material would be severely limited; however it emphasizes the importance of appropriate personalized stimuli selection. From the confusion matrix of these tables, we see that the largest error corresponds to the least populated classes, which is expected. We note that for the case of disgust and contentment we have underrepresented data, which leads to greater error, as random forests favor the majority classes. An additional factor was the subject's unconscious movement as a response to strong stimulus. We should also mention that we only take into consideration the main part of each emotion, excluding any initial transitional samples; i.e. the first 10-30sec of each emotion.

Table V: Results on highest-ranked parts of experiment for 6 emotions problem

| Label | Pred / Real | 1 | 2 | 3 | 4 | 5 | 6 | Class. Error (%) |
|---|---|---|---|---|---|---|---|---|
| Fear | 1 | 71 | 8 | 10 | 0 | 13 | 3 | 32.3 |
| Sadness/Anger | 2 | 15 | 54 | 4 | 2 | 8 | 1 | 35.7 |
| Awe/Reverence | 3 | 9 | 2 | 96 | 6 | 9 | 2 | 22.6 |
| Disgust | 4 | 2 | 1 | 9 | 30 | 5 | 3 | 40.0 |
| Joy/Amusement | 5 | 8 | 1 | 8 | 1 | 105 | 2 | 16.0 |
| Contentment | 6 | 9 | 2 | 15 | 1 | 2 | 45 | 39.2 |
| Average OOB | | - | | | | | | 28.7 |

Table VI: Results on highest-ranked parts of experiment for binary setup

| Real / Pred | Negative | Positive | Class. Error (%) |
|---|---|---|---|
| Negative | 190 | 49 | 20.5 |
| Positive | 37 | 286 | 11.5 |
| Average OOB | - | - | 15.3 |

Finally, with respect to the selection of window size, we see a summary of the results with RF classification on different window sizes, in Table VII. Clear deterioration was observed outside this range. From these results, we conclude that a window of 32 samples, would minimize the prediction error.

Table VII: effect of window size on prediction error

| Window Size | Average OOB (%) | |
|---|---|---|
| | 6 emotions | Binary |
| 16 | 30.2 | 15.5 |
| 32 | 28.7 | 15.3 |
| 64 | 36.0 | 18.0 |

### 4.c.ii Comparison of Classifiers

Random Forests outperform the other classifiers we tried on predicting whether the subject experiences positive or negative feelings (binary setup). A summary of the results for the binary problem can be seen in Table VIII. The reported results are generated using 10-fold cross validation. The mean cross-validation error is reported. As the data are shuffled, and different samples are left out each time to be used as test-sets, we can compare directly the cross-validation error to the OOB error, since both are unbiased estimates of the prediction error. The other classifiers we report on are decision trees, artificial neural networks (ANN) and support vector machines (SVM). The ANN consists of one hidden layer with 10 neurons. It was trained with the back-propagation algorithm and radial basis functions.

The reported results for the SVM classifier are derived after the model parameters were optimized, which led to choosing gamma 0.1 and cost function equal to 10. We see that Random Forests outperform the other classifiers, followed by the SVM classifier, on the binary setup. Additionally, for the 6-emotion problem, the prediction error for SVMs was found to be 33.6% compared to 28.7% for RF.

Table VIII: Comparison of classifiers on the binary setup

| Classifier | Prediction Error (%) |
|---|---|
| Decision Tree | 23.6 |
| Artificial Neural Network | 25.0 |
| Support Vector Machines | 19.0 |
| Random Forests | 15.3 |

**4.d Variable Importance**

The most important features are derived based on Gini index, which is a measure of node impurity, with RF classification. The results are shown in Table IX.

Table IX: Variable importance on 6 emotions

| Signal | Features | 6-emotion Ranking | Binary Ranking |
|---|---|---|---|
| HRV | Mean | 1 | 1 |
| HR | St. Deviation | 2 | 7 |
| GSR | St. Deviation | 3 | 5 |
| HR | Mean | 4 | 4 |
| HRP | Mean | 5 | 3 |
| GSR | Mean | 6 | 9 |
| HRP | St. Deviation | 7 | 2 |
| GSR | Sum of Sq. | 8 | 8 |
| BR | St. Deviation | 9 | 6 |
| BR | Mean | 10 | 11 |
| HRV | Std. of Diff. Sq. | 11 | 12 |
| HRV | Mean of Diff. Sq. | 12 | 13 |
| BR | Sum of Sq. | 13 | 10 |
| HRV | St. Deviation | 14 | 14 |
| HRV | Mean of Diff. | 15 | 15 |
| HRV | Mean of Diff. Sq. | 16 | 16 |

The features are ranked from most to least significant feature; with 1 denoting the most significant feature and 16 the least significant one. From Table IX, we can see that the HR-related features dominate as the most important ones. For both the 6-emotion and the binary setup, the mean of HRV is the most important feature. Mean HRV is followed by the standard deviation of HR and the standard deviation of HRP, as the second most important feature for the 6-emotion and the binary problem, respectively. Out of the top five significant features, for both setups, four are HR-based.

## 5. Summary

The entire pipeline of a controlled setup experiment was presented to set the baseline and train a model at personalized level, for future prediction of the emotional state of the subject in real-life, under static conditions. Main components of this flow are the following: triggers (material selection), protocol, data collection and data analysis (pre-processing, feature extraction, classification and prediction). Random Forest was found to be the more accurate classifier, among the four ones we trained. We have very promising results on the six-emotion experiment, using only easy-to-implement in wearables, time-domain features. The accuracy for the binary case is ~85%. Furthermore, the biosignals were ranked in terms of their effect in classification accuracy and the Heart-Rate and its derivatives were found to be the most significant ones. Finally, an adaptive stimuli selection and controlled-setup session-generation system was proposed to allow for greater accuracy in baseline estimation for emotion recognition. Starting from a large initial stimuli selection, the system could potentially generate the optimal subset of stimuli material, based on the user's feedback, to establish emotion baselines at a personalized level.

## 6. Acknowledgements

The authors would like to thank R. Kumar and his team for providing the internal physiological sensor recording device, as well as G. Takos for his help and support.